\documentclass[runningheads]{llncs}
\usepackage[T1]{fontenc}
\usepackage{graphicx,verbatim}
\usepackage{booktabs} 
\usepackage{multirow} 
\usepackage{array}    
\usepackage{ulem}     
\usepackage{amssymb} 
\usepackage{xcolor}  
\usepackage[table]{xcolor} 
\definecolor{lightgray}{gray}{0.9} 
\usepackage{enumitem}
\usepackage{microtype}
\usepackage{cite}
\begin{document}
\title{DINO-Med3D: Bridging Dimension and Domain Gaps in Volumetric Segmentation via Progressive Adaptation\thanks{Accepted at MICCAI 2026. The camera-ready version and link will be made available upon publication.}}
\titlerunning{DINO-Med3D: Bridging Dimension and Domain Gaps}
%

\author{Haoyu Hu\inst{1,2} \and
Xiyao Ma\inst{1,2} \and
Shiqi Liu\inst{2}\textsuperscript{**}\and
Linsen Zhang\inst{2} \and
Xiaoliang Xie\inst{2} \and
Xiaohu Zhou\inst{2} \and
Zeng-Guang Hou\inst{2}\thanks{Corresponding author}}

\authorrunning{H. Hu et al.}

\institute{University of Chinese Academy of Sciences, Beijing, China \and
Institute of Automation, Chinese Academy of Sciences, Beijing, China \\
\email{\{liushiqi2016, zengguang.hou\}@ia.ac.cn}}

\maketitle              
\begin{abstract}
Although DINOv3 has demonstrated remarkable semantic discrimination in natural imagery, its direct application to volumetric medical segmentation is hindered by inherent dimension and domain disparities. To resolve these issues, we propose DINO-Med3D, a two-stage progressive framework that repurpose the pre-trained DINOv3 encoder for 3D medical tasks. In the first stage, we mitigate the dimension gap by introducing a multi-slice embedding module that incorporates pseudo-3D context, while simultaneously employing a segmentation proxy task to adapt representations learned from natural scenes to the medical domain.
Subsequently, we further enhance volumetric understanding by adding lightweight 3D adapters into the frozen backbone to enforce global inter-slice continuity. 
Finally, to compensate for the spatial information loss inherent in the embedding process, we design a parallel Detail-Recovery stream to explicitly preserve high-frequency boundary cues.
Extensive experiments on five public datasets demonstrate that our approach successfully adapts DINOv3 to the medical domain and significantly outperforms state-of-the-art baselines. The source code is available at https://github.com/HaoyuHu1/DINO-Med3D/.


\end{abstract}
\section{Introduction}
Medical image segmentation plays a pivotal role in computer-aided diagnosis by facilitating quantitative clinical assessment. Despite the architectural evolution from CNNs \cite{ronneberger2015u,cciccek20163d,zhou2018unet++,chen2021transunet,isensee2021nnu} to recent Transformers and Mamba-based models \cite{hatamizadeh2021swin, U-Mamba}, performance remains constrained by the scarcity of high-quality annotations. Models trained on isolated, small-scale datasets often struggle with fully leveraging their potentials.

Foundation models offer a promising avenue to circumvent this data bottleneck. While the recent Segment Anything paradigm \cite{kirillov2023segment} and its medical adaptations \cite{jiang2026medical,liu2025medsam3} demonstrate impressive zero-shot capabilities, they typically rely on interactive prompts or heuristic auto-prompting mechanisms. In contrast, self-supervised discriminative models, exemplified by DINOv3 \cite{caron2021emerging,oquab2023dinov2,simeoni2025dinov3}, learn robust semantic representations directly from data without requiring user interaction. Following this paradigm, a surge of recent studies has adapted these general-purpose encoders to the medical domain \cite{yang2025segdino,scholz2025mm,gao2025dino}.
However, directly repurposing 2D natural image encoders for 3D medical volumes is non-trivial. As highlighted by recent benchmarks \cite{liu2025does}, DINOv3 performance degrades significantly in tasks involving substantial domain shifts or dense 3D predictions, often lagging behind specialized frameworks like nnU-Net \cite{isensee2021nnu}. This suggests that naive fine-tuning or simple projection is insufficient to bridge the feature gap between natural imagery and complex volumetric medical data.

We attribute this performance degradation to two primary disparities: the dimension gap and the domain gap. First, adapting 2D encoders to 3D volumes forces slice-wise processing, inevitably discarding spatial correlations along the z-axis. Second, DINOv3 representations are biased toward the distinct shapes and high contrast of natural scenes, generalizing poorly to medical tasks that rely on subtle texture cues to define low-contrast boundaries.

To systematically bridge these gaps, we propose DINO-Med3D, a progressive framework that evolves the 2D foundation model into a 3D medical segmentor through a two-stage paradigm. In the first stage, we address the initial disparities by designing a specialized multi-slice embedding module, which aggregates contextual information from adjacent slices. Simultaneously, we mitigate the domain gap by employing a segmentation proxy task to adapt fundamental representations learned from natural scenes to medical domain.
In the second stage, we freeze the aligned backbone and introduce lightweight 3D adapters alongside Low-Rank Adaptation (LoRA) \cite{hu2022lora} to enforce global inter-slice continuity, thereby further bridging the dimension gap. Concurrently, a parallel Detail-Recovery stream is designed to explicitly capture high-frequency boundary cues, facilitating precise adaptation to the textural properties of medical images. Our contributions are summarized as follows: 
\begin{itemize}[label={$\bullet$}] 
\item We propose DINO-Med3D, a novel architecture that effectively repurposes the 2D DINOv3 for 3D medical segmentation by progressively bridging dimension and domain gaps. 
\item We introduce a decoupled two-stage transfer strategy: the first stage aligns feature distributions and input dimensions, while the second stage fosters volumetric reasoning and texture refinement via 3D adapters and the proposed Detail-Recovery stream. 
\item We conduct extensive experiments on five public datasets spanning multiple modalities, anatomical structures, and pathological textures. The results demonstrate that our method consistently bridges the aforementioned gaps and significantly outperforms SOTA baselines. 
\end{itemize}

\section{Methods}
\subsection{Overall Architecture} As illustrated in Fig. 1, the proposed framework bridges 2D pre-training and 3D volumetric segmentation through a two-stage approach. In Stage I, we adapt the standard 2D DINOv3 encoder to the medical domain by employing a dimension-aware adaptation mechanism alongside a segmentation proxy task. Subsequently, in Stage II, we construct a dual-stream architecture comprising (1) a Volumetric Context Encoding Stream initialized with Stage I weights for high-level semantic extraction; and (2) a lightweight Detail-Recovery Stream to capture high-frequency spatial information. These features are aggregated via a fusion module and decoded to generate the final segmentation masks.
\begin{figure}[!t]
\includegraphics[width=\textwidth]{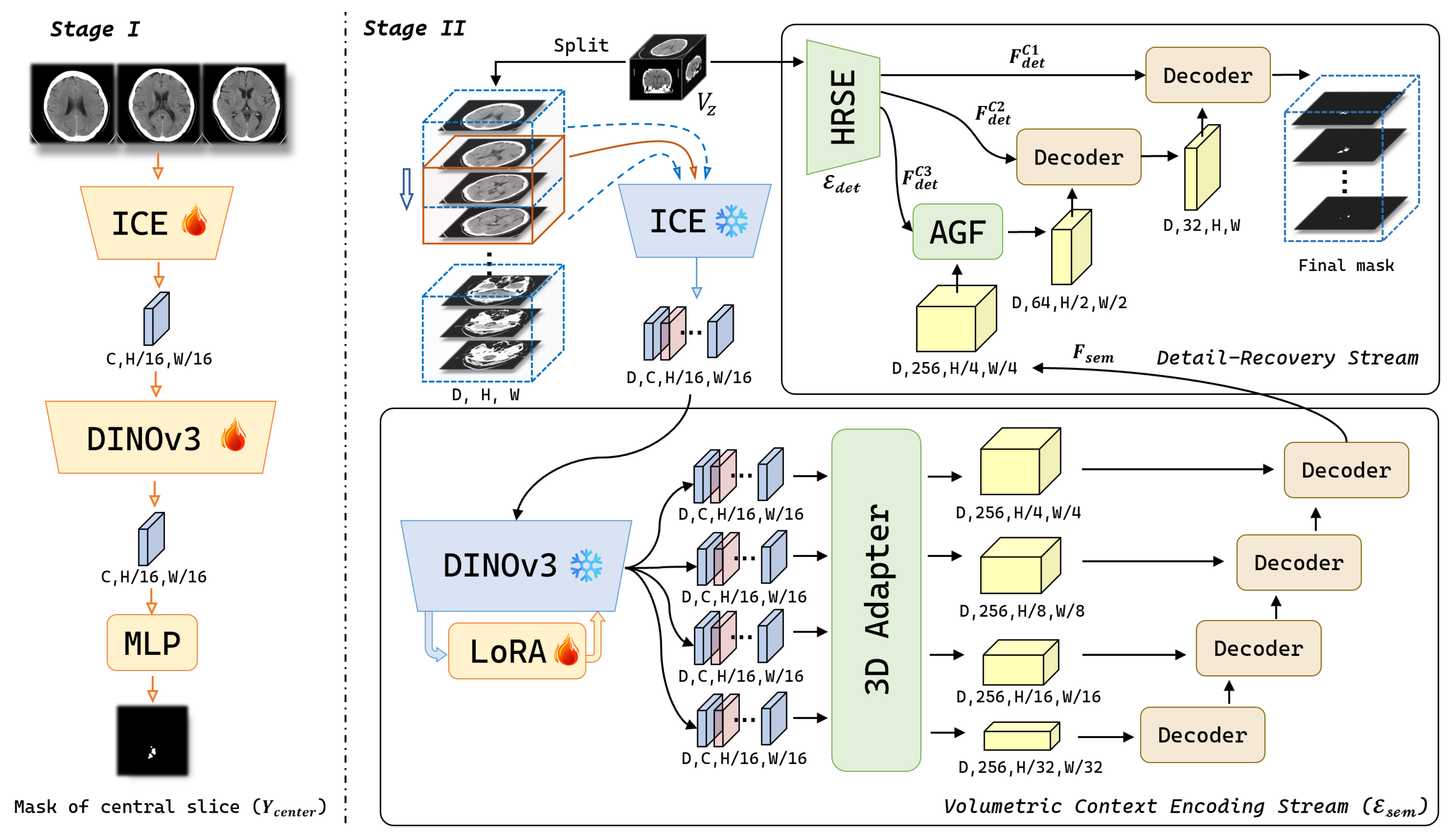}
\caption{Schematic illustration of DINO-Med3D framework. Stage I aligns the pre-trained 2D DINOv3 backbone with volumetric medical data via the Inter-slice Context Embedding (ICE) module and a pseudo-3D proxy task. Stage II freezes the aligned backbone and introduces a dual-stream architecture: a Volumetric Context Encoding Stream leveraging LoRA and lightweight 3D adapters for semantic reasoning, and a parallel Detail-Recovery Stream for capturing high-frequency boundary information. Finally, features are integrated via Adaptive Gated Fusion (AGF) for dense prediction.}
\label{fig:main_pic}
\end{figure}

\subsection{Stage I: Volumetric Adaptation and Alignment}
To bridge the gap between 2D pre-training and volumetric medical data, we introduce a pseudo-3D adaptation strategy facilitated by a depth-aware embedding module and a proxy task.

\subsubsection{Inter-slice Context Embedding(ICE)}Direct 3D input is incompatible with DINOv3's pre-trained 2D Rotary Positional Embeddings, while naive slice-wise processing discards critical through-plane continuity. To resolve this, we propose ICE to replace the original 2D patch embedding. ICE processes a sub-volume slab $V \in \mathbb{R}^{K \times H \times W}$ to yield a pseudo-3D patch embedding of size $C \times \frac{H}{16} \times \frac{W}{16}$ via depth-aware 3D convolution. Here, $C$ denotes the embedding dimension of DINOv3 and $K$ is a tunable hyperparameter representing the input slices of ICE. 
In this way, while maintaining dimensional compatibility with the 2D backbone, ICE effectively fuses inter-slice context and learns volumetric structure.

\subsubsection{Proxy Task: Domain Adaptation}We further align the feature space with medical semantics via a proxy segmentation task. A lightweight MLP head projects the encoder's output from $\mathbb{R}^{C \times \frac{H}{16} \times \frac{W}{16}}$ to $\mathbb{R}^{H \times W}$ to predict the segmentation map $\hat{Y}_{center}$ of the central slice. The ICE module and backbone are fully fine-tuned to minimize the discrepancy between $\hat{Y}_{center}$ and the ground truth, compelling the model to adapt feature representations from natural scenes to medical images.

\subsection{Stage II: Dual-Stream Segmentation Network}
To synergize semantic abstraction with spatial precision, we propose a dual-stream architecture. Leveraging the Stage I initialization, the framework processes input $X \in \mathbb{R}^{D \times H \times W}$ to generate a spatially corresponding dense prediction map $M$, where $D$ represents the input depth. Two parallel paths are constructed: a context stream $\mathcal{E}_{sem}$ for long-range semantic features ($F_{sem}$) and a lightweight detail stream $\mathcal{E}_{det}$ for high-frequency features ($F_{det}$). These hierarchical features are integrated via a fusion module $\mathcal{F}$ and reconstructed by decoder $\mathcal{D}$ to yield $M$:
\begin{equation}
M = \mathcal{D}\left( \mathcal{F}(F_{sem}, F_{det}), F_{det}\right) = \mathcal{D}\left(  \mathcal{F}(\mathcal{E}_{sem}(X), \mathcal{E}_{det}(X)), \mathcal{E}_{det}(X)\right)
\end{equation}
\subsubsection{Volumetric Context Encoding Stream}
This stream($\mathcal{E}_{sem}$) leverages the representations from Stage I and extends these capabilities to full volume. Given an input volume $X \in \mathbb{R}^{D \times H \times W}$, we construct a sequence of pseudo-3D slabs $V$. For depth $z$=1,2,...,D, adjacent $K$ slices are stacked to form a multi-channel slab $V_z = [x_{z-\frac{K-1}{2}}, \dots, x_z, \dots, x_{z+\frac{K-1}{2}}] \in \mathbb{R}^{K \times H \times W}$, utilizing zero-padding at boundaries. These D slabs are processed by the ICE and backbone sequentially and then stacked to restore volumetric geometry. To preserve the 3D medical representation acquired in stage I, we freeze the pre-trained backbone and inject trainable LoRA matrices into the Transformer blocks. A lightweight 3D Adapter aligns hierarchical features via 3D deconvolutions to transition from the stacked pseudo-3D features to genuine volumetric
representation, followed by a UperNet \cite{xiao2018unified} decoder to yield the coarse semantic map $F_{sem} \in \mathbb{R}^{D \times C_{sem} \times \frac{H}{4} \times \frac{W}{4}}$.
\subsubsection{Detail-Recovery Stream}
Complementing the semantic stream, the High-Resolution Spatial Encode (HRSE, $\mathcal{E}_{det}$) compensates for the loss of high-frequency spatial details inherent in patch-based embedding. Designed as a lightweight 3D CNN with residual blocks, HRSE prioritizes spatial precision by avoiding aggressive downsampling, extracting a feature hierarchy $F_{det} = \{C_1, C_2, C_3\}$ at original, $1/2$, and $1/4$ scales. This preserves fine-grained boundary information essential for segmentation.
To bridge the semantic-spatial gap between two streams, we introduce the Adaptive Gated Fusion (AGF, $\mathcal{F}$) module, which acts as a content-aware residual correction mechanism. Given the semantic features $F_{sem}$ and local details $F_{det}^{C3}$, AGF first generates a spatial attention map $G$ to highlight regions requiring refinement:
\begin{equation}
G = \sigma\left( \mathcal{K}_{1\times1\times1}\left( [F_{sem}; F_{det}^{C3}] \right) \right)
\end{equation}
where $[\;;\,]$ denotes channel-wise concatenation, $\mathcal{K}_{1\times1\times1}$ is a $1\times1\times1$ convolution and $\sigma$ is the sigmoid function. The gate $G \in [0, 1]$ then modulates the injection of high-frequency details into the semantic stream:
\begin{equation}
\tilde{F}_{det}^{C3} = F_{sem} + G \odot F_{det}^{C3}
\end{equation}
Finally, the refined $\tilde{F}_{det}^{C3}$ is progressively aggregated with lower-level features ($F_{det}^{C1, C2}$) via a UperNet decoder to yield the final probability map.

\section{Experiments and Results}
\subsection{Dataset}
We evaluated our method on five public datasets across CT and MRI modalities. CT datasets include: (1) AISD \cite{liang2021symmetry}: 397 brain scans with infarct lesions; (2) MSD-Pancreas and (3) MSD-Colon \cite{antonelli2022medical}: 281 and 126 abdominal scans, cropped to 50 and 30 slices along the z-axis, respectively. MRI datasets include: (4) BraTS 2020 \cite{menze2014multimodal , bakas2017advancing , bakas2018identifying}: 368 brain scans (FLAIR, Whole Tumor), downsampled to 4 mm spacing; and (5) ACDC \cite{bernard2018deep}: 100 cardiac scans. All images were resized to 512$\times$512 and randomly split (8:1:1).
Specific windowing (WW, WL) was applied for CTs: AISD (100, 40), Pancreas (255, 50), and Colon (400, 40). MRIs were clipped to the [0.5, 99.5] percentile range and normalized.

\subsection{Implementation Details and Evaluation Metrics}
Standard data augmentation (rotation, scaling, intensity jitter) was employed. We optimized a compound loss $\mathcal{L} = 0.2\mathcal{L}_{CE} + 0.8\mathcal{L}_{Dice}$ for both stage (50 epochs each). Stage 1: batch size (bs) 24, learning rate (lr) $2 \times 10^{-5}$. Stage 2: initialized from the best Stage 1 checkpoint, fine-tuned with LoRA ($r=16, \alpha=16, p=0.1$), bs 1, and lr $1 \times 10^{-4}$. The ICE input slices \textit{K} is set to 3. Our proposed method variants are denoted as DINO-Med3D-S/B/L. All baseline methods were implemented following their default configurations. All models were trained for 100 epochs. Evaluation metrics include Dice Similarity Coefficient (DSC) and Hausdorff Distance (HD95).
\subsection{Result}
\begin{table}[t]
    \centering
    \caption{Quantitative comparison. The best baselines are in \colorbox{lightgray}{gray}. Scores exceeding the best and second-best baseline are \textbf{bolded} and \underline{underlined} respectively.}
    \label{tab:main_results}
    
    \fontsize{8}{9.5}\selectfont 
    
    \setlength{\tabcolsep}{1.1pt} 
    
    \begin{tabular}{l cc|cc|cc|cc|cc c}
        \toprule
        
        \multirow{2}{*}{\textbf{Method}} & \multicolumn{2}{c}{AISD} & \multicolumn{2}{c}{MSD-Pan.} & \multicolumn{2}{c}{MSD-Colon} & \multicolumn{2}{c}{BraTS} & \multicolumn{2}{c}{ACDC} & \multirow{2}{*}{\shortstack{Params\\(M)}} \\
        
        \cmidrule(lr){2-3} \cmidrule(lr){4-5} \cmidrule(lr){6-7} \cmidrule(lr){8-9} \cmidrule(lr){10-11}
        
        & DSC$\uparrow$ & \multicolumn{1}{c}{HD95$\downarrow$} 
        & DSC$\uparrow$ & \multicolumn{1}{c}{HD95$\downarrow$} 
        & DSC$\uparrow$ & \multicolumn{1}{c}{HD95$\downarrow$} 
        & DSC$\uparrow$ & \multicolumn{1}{c}{HD95$\downarrow$} 
        & DSC$\uparrow$ & HD95$\downarrow$ & \\ 
        \midrule
        
        nnU-Net    & 47.44 & 57.09 & \cellcolor{lightgray}{56.93} & \cellcolor{lightgray}{6.93}  & 38.13 & \cellcolor{lightgray}{34.44} & 88.73 & 3.62 & 88.63 & 1.57 & 31 \\
        nnFormer   & 34.87 & 107.75& 54.42 & 10.83 & \cellcolor{lightgray}{50.68} & 68.91 & 87.31 & 24.40& 75.77 & 6.03 & 150 \\
        SwinUNETR  & 42.73 & 45.91 & 51.85 & 20.51 & 20.84 & 79.12 & 88.76 & 7.38 & 85.96 & 3.64 & 62 \\
        Umamba     & 41.92 & 62.82 & 53.00 & 14.62 & 32.33 & 64.20 & \cellcolor{lightgray}{89.61} & \cellcolor{lightgray}{3.30} & 87.70 & 1.32 & 43 \\
        VoCo(Swin-L)  & 43.99 & \cellcolor{lightgray}{39.26} & 51.37 & 33.72 & 33.51 & 60.68 & 89.19 & 9.26 & 85.57 & 6.56 & 248 \\
        Dino U-Net  & \cellcolor{lightgray}{48.19} & 56.79 & 48.25 & 14.01 & 31.26 & 91.51 & 89.09 & 5.38 & \cellcolor{lightgray}{91.19} & \cellcolor{lightgray}{1.17} & 318 \\
        \midrule 
        
        our-Small & \textbf{51.86} & 51.76 & \textbf{58.58} & \underline{9.36} & \underline{47.08} & \underline{50.36} & \textbf{89.96} & \textbf{2.18} & \underline{88.81} & \textbf{1.12} & 47 \\
        our-Base  & \textbf{52.72} & 49.75 & \textbf{59.50} & \underline{7.08}  & \textbf{53.28} & \underline{34.61} & \textbf{90.44} & \textbf{2.34} & \underline{90.82} & \textbf{0.91} & 112 \\
        our-Large & \textbf{53.85} & \underline{44.23} & \textbf{61.20} & \textbf{6.40}  & \textbf{62.28} & \textbf{27.70} & \textbf{90.27} & \textbf{2.06} & \textbf{91.52} & \textbf{0.86} & 334 \\
        
        \bottomrule
    \end{tabular}
\end{table}
\begin{figure}[!ht]
\includegraphics[width=\textwidth]{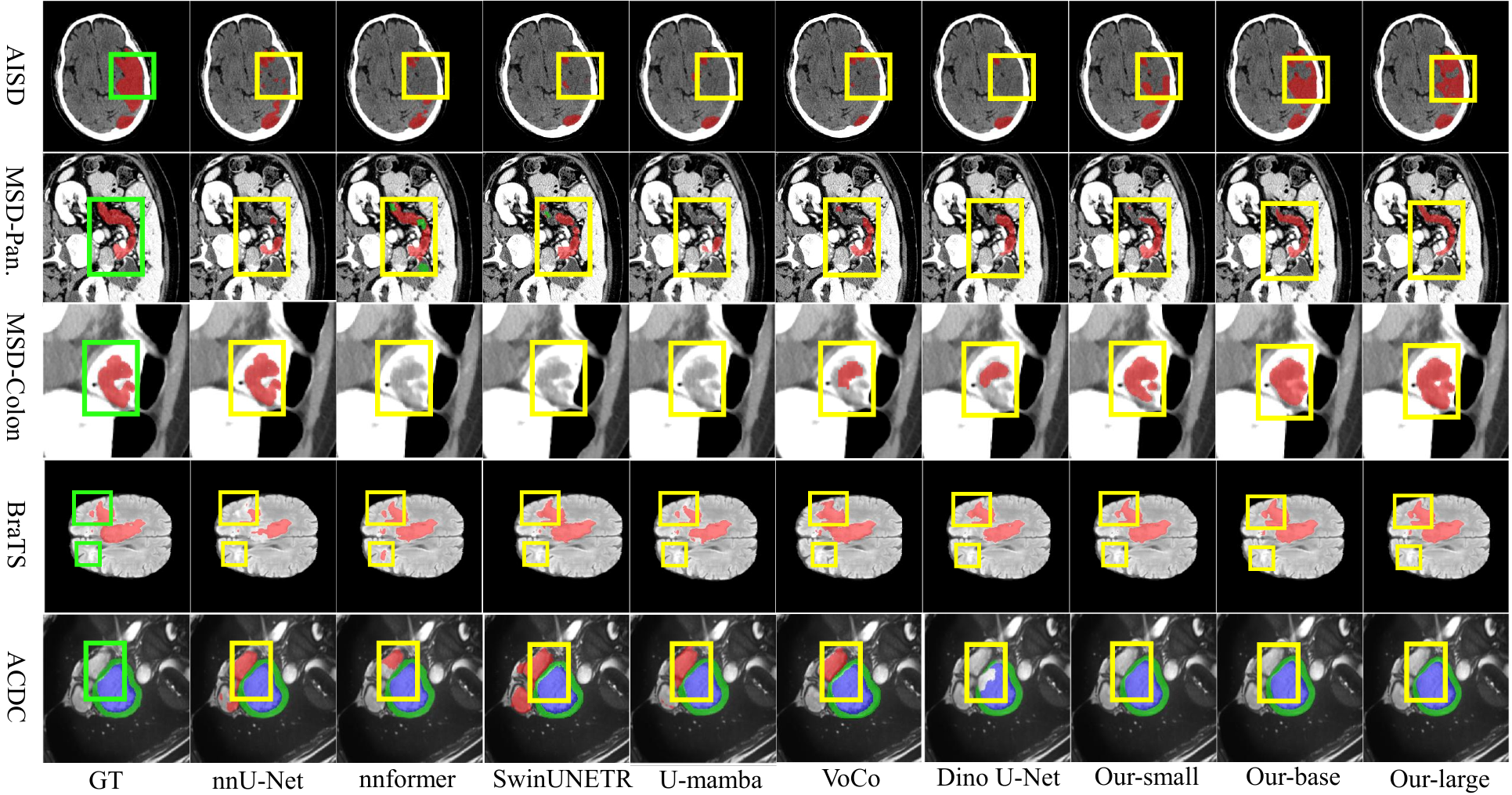}
\caption{Qualitative comparison on representative samples. Images are cropped and zoomed for better visualization. Prominent regions are highlighted with boxes.} \label{fig:main_result}
\end{figure}

Table \ref{tab:main_results} presents the quantitative comparisons across different benchmarks and Figure \ref{fig:main_result} provides a qualitative comparison. Our proposed framework consistently outperforms SOTA baselines, including the CNN-based nnU-Net \cite{isensee2021nnu}, the Transformer-based nnFormer \cite{zhou2023nnformer} and SwinUNETR \cite{hatamizadeh2021swin}, and the Mamba-based U-Mamba \cite{U-Mamba}. Furthermore, we benchmark our method against VoCo \cite{wu2025large}, a representative self-supervised pre-training framework. We also include Dino U-Net \cite{gao2025dino}, a representative DINOv3-based adaptation method for medical imaging. Notably, the DINO-Med3D-L variant achieves the best overall performance in terms of both DSC and HD95. This demonstrates the framework's robustness and potent segmentation capabilities across multiple modalities (CT and MRI), diverse anatomical structures, and varying segmentation targets.

Our method demonstrates superior robustness in challenging scenarios characterized by high anatomical variability, such as the MSD-Colon and AISD datasets. Notably, on MSD-Colon, DINO-Med3D-L achieves a DSC of 62.28\%, outperforming SOTA method nnFormer (50.68\%) by a substantial margin of 11.60\%. In contrast, competing approaches suffered from severe under-segmentation, further highlighting the efficacy of our framework in handling complex cases.

Direct comparison with Dino U-Net highlights the efficacy of our architectural adaptations. To ensure a fair evaluation, we benchmark our model against the Large variant model of Dino U-Net which is built upon the DINOv3 architecture with large backbones. Despite both methods leveraging DINOv3 encoders, DINO-Med3D yields consistently higher accuracy, particularly on texture-rich datasets (MSD-Panc. and Colon). Furthermore, we observe a positive scaling law: performance improves monotonically from Small to Large variants (Params: 47M to 334M). It is worth noting that even our efficiency-focused Small variant (47M) outperforms the much larger Dino U-Net (318M) on AISD, MSD and BraTS, striking a favorable trade-off between computational cost and segmentation accuracy.

\subsection{Ablation Study}
To validate the efficacy of the proposed components, we conducted ablation studies on three datasets—MSD-Pancreas, MSD-Colon, and BraTS—covering diverse organs and modalities. All implementations were built upon the DINO-Med3D-B.

Table \ref{tab:ablation_layers} analyzes the impact of ICE input slices \textit{K}. As presented in Table \ref{tab:ablation_layers}, performance improves as the number of layers increases, peaking at three layers across all datasets, before degrading with further additions. Specifically, on MSD-Pancreas, the DSC rises from 54.97\% to 59.50\% and subsequently drops to 58.72\%. This suggests that while multi-layer encoding facilitates better feature extraction, excessive depth may compromise the model's representation capability.
\begin{table}[htbp]
    \centering
    \caption{Impact of the ICE input slices \textit{K} on segmentation performance. Results are reported in DSC. Best scores are bolded. }
    \label{tab:ablation_layers}
    
    \fontsize{8}{9.5}\selectfont 
    \setlength{\tabcolsep}{3.5pt} 
    
    \begin{tabular}{l ccccc}
        \toprule
        \multirow{2}{*}{Dataset} & \multicolumn{5}{c}{ICE input slices \textit{K}} \\
        \cmidrule(lr){2-6}
         & 1 & 2 & 3 & 4 & 5 \\
        \midrule
        
        MSD-Pan. & 
        54.97  & 
        55.97 & 
        \textbf{59.50} & 
        58.79  & 
        58.72  \\
        
        MSD-Colon & 
        44.26 & 
        51.99 & 
        \textbf{53.28} & 
        47.53  & 
        33.10  \\
        
        BraTS & 
        89.66  & 
        90.01  & 
        \textbf{90.44} & 
        89.69  & 
        90.20  \\
        
        \bottomrule
    \end{tabular}
\end{table}

\begin{figure}[ht]
\includegraphics[width=\textwidth]{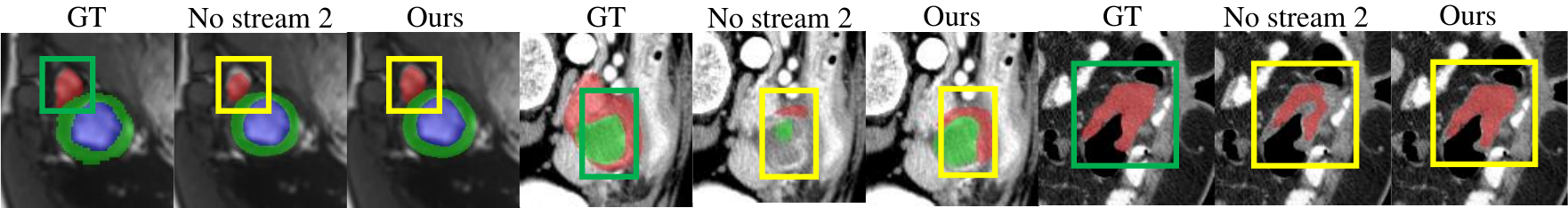}
\caption{Visual ablation study of the Detail-Recovery Stream. Cases from ACDC and MSD (Panc. \& Colon) are presented. Each triplet shows: (a) Ground Truth, (b) w/o Stream 2, and (c) Full Method. Boxes indicate major improvements in detail preservation.} \label{fig:no_res}
\end{figure}

\begin{table}[htbp]
    \centering
    \caption{Ablation study results. The bolded row represents the full implementation of our proposed method.}
    \label{tab:ablation}
    
    \fontsize{8}{9.5}\selectfont 
    \setlength{\tabcolsep}{2pt} 
    
    \begin{tabular}{cc cc ccc}
        \toprule
        \multicolumn{2}{c}{Stage I} & \multicolumn{2}{c}{Stage II} & \multicolumn{3}{c}{Dice Similarity Coefficient $\uparrow$} \\
        \cmidrule(lr){1-2} \cmidrule(lr){3-4} \cmidrule(lr){5-7}
        
        UperNet & MLP & Stream 1 & Stream 2 & MSD-Pan. & MSD-Colon & BraTS \\
        \midrule
        
         & & & & 
        28.60 \scriptsize{(-30.90)} & 17.38 \scriptsize{(-35.90)} & 78.46 \scriptsize{(-11.98)} \\
        
         & \checkmark & & & 
        54.63 \scriptsize{(-4.87)} & 43.11 \scriptsize{(-10.17)} & 89.16 \scriptsize{(-1.28)} \\
        
         & \checkmark& \checkmark & & 
        57.00 \scriptsize{(-2.50)} & 44.93 \scriptsize{(-8.35)} & 89.91 \scriptsize{(-0.53)} \\
        
         & \checkmark& \checkmark & \checkmark & 
        \textbf{59.50} & \textbf{53.28} & \textbf{90.44} \\
        
         &  & \checkmark & \checkmark & 
        54.78 \scriptsize{(-4.72)} & 45.96 \scriptsize{(-7.32)} & 90.19 \scriptsize{(-0.25)} \\
        
        \checkmark & &\checkmark & \checkmark & 
        57.42 \scriptsize{(-2.08)} & 47.95 \scriptsize{(-5.33)} & 90.39 \scriptsize{(-0.05)} \\
        
        \bottomrule
    \end{tabular}
\end{table}
Table \ref{tab:ablation} dissects the efficacy of the proposed multi-stage framework. A frozen DINOv3 backbone with a direct MLP decoder yields suboptimal results (Row 1). Conversely, activating Stage I—unfreezing the backbone via the proxy segmentation task with ICE module—yields significant gains. In Stage II, introducing Stream 1 (Volumetric Context Encoding) boosts MSD-Pancreas DSC by 2.37\%, confirming the necessity of transforming pseudo-3D features into intrinsic 3D representations. Integrating Stream 2 (Detail-Recovery) further refines boundaries, with qualitative improvements shown in Fig. \ref{fig:no_res}.

Finally, we verified the role of the proxy task and the decoder architecture. Removing Stage I leads to a 4.72\% DSC drop on MSD-Pancreas. Notably, this drop occurs despite the application of LoRA in Stage II. This demonstrates that Stage I's full fine-tuning is a prerequisite for robust transfer learning, as the Parameter-Efficient LoRA adaptation in Stage II cannot fully bridge for the dimension and domain gaps. While retaining Stage I and substituting the MLP decoder with a UperNet decoder improves performance compared to the no-Stage-I setting (+2.64\% DSC), it fails to match the optimal results, thereby demonstrating the effectiveness of our lightweight decoder in proxy task.

\section{Conclusion and Discussion}
In this work, we proposed DINO-Med3D, a progressive adaptation framework designed to repurpose the 2D DINOv3 foundation model for volumetric medical segmentation. Comprehensive evaluations across five datasets demonstrate that DINO-Med3D consistently outperforms state-of-the-art baselines, validating the efficacy of our dual-stream architecture.

Despite these advancements, certain limitations warrant further investigation. First, the performance variability of the Detail-Recovery Stream across datasets suggests a dependency on texture complexity and lesion size that warrants deeper investigation. Second, the first stage necessitates full fine-tuning of the backbone to align feature distributions, imposing a substantial computational burden. Future work will focus on developing more parameter-efficient alignment strategies.

%
%
%
\bibliographystyle{splncs04}
\bibliography{Paper-3353}
%




\end{document}